\documentclass[sn-mathphys-num]{sn-jnl}

\usepackage{graphicx}%
\usepackage{multirow}%
\usepackage{amsmath,amssymb,amsfonts}%
\usepackage{amsthm}%
\usepackage{mathrsfs}%
\usepackage[title]{appendix}%
\usepackage{xcolor}%
\usepackage{textcomp}%
\usepackage{manyfoot}%
\usepackage{booktabs}%
\usepackage{algorithm}%
\usepackage{algorithmicx}%
\usepackage{algpseudocode}%
\usepackage{listings}%
\usepackage{graphicx}
\usepackage{subcaption}
\usepackage{graphicx}
\usepackage{booktabs}
\usepackage{xcolor}
\usepackage{multirow}
\usepackage{adjustbox}
\usepackage{tabularx}
\usepackage{float}  
\usepackage{afterpage}
\usepackage{calc}
\usepackage{booktabs}
\usepackage{tabularx}
\usepackage{threeparttable}

\definecolor{darkgreen}{rgb}{0,0.6,0}

\theoremstyle{thmstyleone}%
\theoremstyle{thmstyletwo}%

\theoremstyle{thmstylethree}%

\begin{document}

\title[Article Title]{Clustering-based Feature Representation Learning for Oracle Bone Inscriptions Detection}


\author[1]{\fnm{Ye} \sur{Tao}}\email{taoye23@mails.jlu.edu.cn}

\author[1]{\fnm{Xinran} \sur{Fu}}\email{fuxr21@mails.jlu.edu.cn}
\author[1]{\fnm{Honglin} \sur{Pang}}\email{panghl22@mails.jlu.edu.cn}
\author*[1,3,4]{\fnm{Xi} \sur{Yang}}\email{yangxi21@jlu.edu.cn}
\author*[2,4]{\fnm{Chuntao} \sur{Li}}\email{lct33@jlu.edu.cn}

\affil[1]{\orgdiv{School of Artificial Intelligence}, \orgname{Jilin University}, \orgaddress{\street{Qianjin Street}, \city{Changchun}, \postcode{130000}, \state{Jilin}, \country{China}}}

\affil[2]{\orgdiv{School of Archaeology}, \orgname{Jilin University}, \orgaddress{\street{Qianjin Street}, \city{Changchun}, \postcode{130000}, \state{Jilin}, \country{China}}}


\affil[3]{\orgdiv{Engineering Research Center of Knowledge-Driven Human-Machine Intelligence}, \orgname{MoE}, \orgaddress{ \city{Changchun}, \postcode{130000}, \state{Jilin}, \country{China}}}
\affil[4]{\orgdiv{Key Laboratory of Ancient Chinese Script, Cultural Relics and Artificial Intelligence}, \orgname{Jilin University}, \orgaddress{ \city{Changchun}, \postcode{130000}, \state{Jilin}, \country{China}}}



\abstract{
Oracle Bone Inscriptions (OBIs), play a crucial role in understanding ancient Chinese civilization. The automated detection of OBIs from rubbing images represents a fundamental yet challenging task in digital archaeology, primarily due to various degradation factors including noise and cracks that limit the effectiveness of conventional detection networks.
To address these challenges, we propose a novel clustering-based feature space representation learning method. Our approach uniquely leverages the Oracle Bones Character (OBC) font library dataset as prior knowledge to enhance feature extraction in the detection network through clustering-based representation learning. The method incorporates a specialized loss function derived from clustering results to optimize feature representation, which is then integrated into the total network loss.
We validate the effectiveness of our method by conducting experiments on two OBIs detection dataset using three mainstream detection frameworks: Faster R-CNN, DETR, and Sparse R-CNN. Through extensive experimentation, all frameworks demonstrate significant performance improvements.

}

\maketitle

\section{Introduction}\label{sec1}

{Oracle Bone Inscriptions (OBIs), as the earliest mature writing system in East Asia, represent an invaluable cultural heritage that provides crucial insights into ancient Chinese civilization. Rubbing images, created by placing paper over raised, incised, or textured surfaces and rubbing it with a colored substance, have been used to preserve and study these inscriptions without damaging the original artifacts. Hence the automated detection of OBIs from rubbing images serves as a fundamental prerequisite for character decoding and subsequent historical research}, contributing significantly to various fields including traditional Chinese philosophy, astronomy, calendar studies, and historical geography. While traditional methods relied heavily on manual expertise, which was both time-consuming and resource-intensive, modern deep learning approaches have revolutionized this process by offering more efficient and cost-effective solutions.{~\cite{nodeep1,nodeep2,nodeep3}}
However, the detection of OBIs from rubbing images presents several unique challenges: (1) The rubbing images are often contaminated with substantial background noise. (2) The presence of cracks in the rubbing images, which share similar textural properties with OBIs, makes differentiation particularly challenging.

\begin{figure}
\centering
\includegraphics[width=1.0\textwidth]{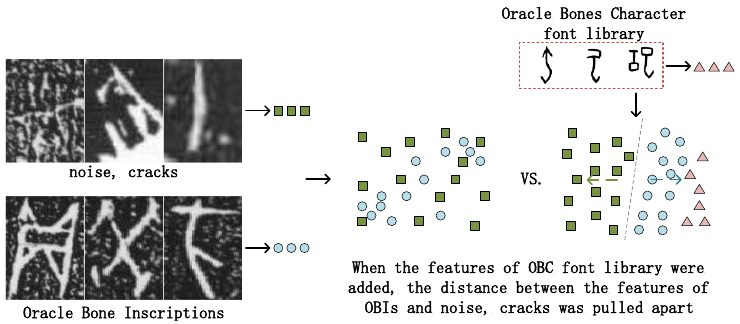}      
\caption{Idea of our proposed method. The features of OBIs and noise, cracks were originally mixed together, and when the features of OBC font library were added, the distance between them was pulled apart.
}
\label{fig:intro} 
\end{figure}

To address these limitations, we propose a novel clustering-based feature space representation learning method that leverages the Oracle Bones Character (OBC) font library dataset~\cite{zitiku} as prior knowledge, {as shown in Fig. \ref{fig:intro}.} Our approach is founded on the intuition that OBIs and non-OBIs (such as cracks and noise) should occupy distinct regions in the deep feature space. By utilizing the OBC font library as a clean, expert-curated reference point, we guide the model to learn more discriminative feature representations that better distinguish between authentic characters and artifacts.

{Previous research in this domain has primarily developed along three key trajectories. The first line of work focuses on OBIs detection.}

Previous work has focused on applying deep learning networks to OBIs detection and proposing improved modules for OBIs dataset difficulties.
Wang et al.~\cite{OBIsdetect0} proposed a dynamic augmentation algorithm based on font dataset and an identification auxiliary detection algorithm, and improved the accuracy of compound graphs detection. Xing et al.~\cite{OBIsdetect3} used the mainstream object detection models to carry out experiments on an OBIs detection dataset and proposed a {data augmentation} algorithm based on several kinds of noises extracted from the OBIs. Chen et al.~\cite{OBIsdetect-1} proposed a method {where shape-adaptive} Gaussian kernels are employed to represent the spatial regions of different OBIs. Fu et al. proposed two algorithms to improve the performance of OBIs detection, namely feature fusion based on cross-attention mechanism method~\cite{OBIsdetect4} and pseudo-category labels prediction method~\cite{OBIsdetect5}.
The above methods can be mainly divided into two categories. The first is to increase the number of difficult samples such as noise and compound graphs utilizing data augmentation to improve the detection effect on difficult samples. The second type is to improve the detection effect by introducing prior knowledge related to OBIs location information, which is obtained through a trained supervised model in advance. The differences between the methods we proposed and others are as follows: 
(1) The method we proposed is end-to-end, and the introduction of prior knowledge does not need to train other networks in advance. 
(2) {Our method} lengthens the distance between OBIs and non-OBIs at the feature space, which improves the detection model's ability to distinguish them from the data structure, thus improving the detection effect.

{A second significant research stream addresses text detection challenges.} Previous works in scene text detection can be broadly categorized into regression-based and segmentation-based methods.

Regression-based Methods use the anchor to generate lots of candidate boxes, {then NMS\cite{softnms,weightednms} is used to obtain the detection results. Firstly, there are methods inspired by object detection. At this stage, the scene text detection method directly locates text by modifying the region proposal of the object detector and the bounding box regression module.  TextBoxes ~\cite{textboxes} define the default box as a quadrilateral with different aspect ratio specifications and use SSD~\cite{ssd} to adapt to different directions and aspect ratios of the text.
TextBoxes++~\cite{textboxes++} extended horizontal text detection to arbitrary orientations.
EAST~\cite{east} simplified the detection pipeline by directly predicting text coordinates and angles.
Secondly, there are methods based on sub-text components.
CTPN~\cite{CTPN} introduced vertical anchor regression to localize text lines accurately.
SegLink~\cite{seglink} extends CTPN by considering multi-directional links between fragments.
There are other methods to improve specific problems. 
LOMO~\cite{lomo} addressed challenging cases such as long text through iterative refinement, while SBD~\cite{SBD} improved robustness by discretizing quadrilateral boxes into key edges.}

Segmentation-based Methods approach text detection through pixel-level semantic segmentation. 
The Mask-TextSpotter series is An End-to-End Trainable Neural Network for Spotting Text with Arbitrary Shapes. The V1 version\cite{masktext} used Mask R-CNN as baseline, and can perform end-to-end text spotting. Its mask branch can generate the text instance segmentation maps and the character segmentation maps. The V2 version\cite{v2} proposed a spatial attention module to enhance the performance and universality. The V3 version\cite{v3} adopts a Segmentation Proposal Network (SPN) instead of the region proposal network (RPN), which gives accurate representations of arbitrary-shape proposals.
{To address the difficulty when recognize curved, irregular text, CRAFT~\cite{CRAFT} detected individual characters and combined them using affinity scores, but it needs complex training. DBNet~\cite{dbnet} introduced differentiable binarization to adaptively set thresholds, reducing dependency on post-processing.}

{The last key research direction centers on cluster-based representation learning.} Self-supervised learning~\cite{moco,mocov2,mocov3,simclr,simclrv2,swav} means that for unlabeled data, the pretext task is designed to mine its own representation features as supervised information to improve the feature extraction ability of the model. In this field, clustering-based representation learning methods have emerged as a particularly promising approach. 
DEC~\cite{DEC} operates by iteratively optimizing a clustering objective based on KL divergence, using a self-training target distribution.
JULE ~\cite{JULE} integrates agglomerative clustering with CNNs and frames this combination as a recurrent process.
DeepCluster~\cite{uc3} iteratively clusters deep features and utilizes the resulting cluster assignments as pseudo-labels to train the convolutional neural network.
DeeperCluster~\cite{DeeperCluster} is designed to handle large volumes of uncurated data.
Tuo et al. proposed a clustering-based supervised learning scheme for {point cloud analysis~\cite{julei}. They conducted} within-class clustering to learn an appropriate point embedding space that is aware of both discriminative semantics and challenging variations. {In order to solve the problem that autoencoder is sensitive to noise data in multi-view clustering research, Fatemeh Daneshfar et al. proposed an Elastic Deep Multi-view Autoencoder with Diversity Embedding (EDMVAE-DE) method~\cite{elastic}, which has adaptable elastic loss, diversity constraint and Graph regularization to be capable of handling noisy data, it can also make full use of multi-view data.}


Specifically, our method integrates both the OBIs detection dataset~\cite{OBIsdetect2} and the OBC font library dataset during the training process. Through a specialized loss function derived from clustering results, we optimize the feature representation to maximize the separation between true characters and noise while maintaining consistency with the standard forms from the font library. This approach offers several key advantages:
1) It introduces a novel methodology for incorporating expert knowledge through the OBC font library dataset, using pristine character features as anchors for feature space organization.
2) It provides a simple yet effective enhancement that can be readily integrated into existing detection frameworks without requiring complex architectural modifications.
3) It directly addresses the core challenges of OBIs detection by improving the model's ability to discriminate between authentic characters and various forms of interference.

To validate our approach, we conducted extensive experiments using three mainstream detection frameworks: Faster R-CNN~\cite{fasterrcnn}, DETR~\cite{DETR}, and Sparse R-CNN~\cite{sparsercnn}. The results demonstrate consistent performance improvements across all frameworks, with significant gains in detection accuracy and robustness. Through feature visualization and detailed analysis of detection results, we confirm that our method effectively enhances character feature discrimination while maintaining resilience against various forms of interference.
We make several contributions to the field of automated OBIs analysis and digital preservation of ancient writing systems, offering a novel approach that combines traditional expertise with modern deep learning techniques to achieve superior detection performance.

\section{Methods}\label{sec4}
\subsection{Dataset}\label{sec3}
There are the introductions of the datasets used in our experiment.

\textbf{OBIs Detection Dataset.}
We conducted experiments on two datasets. Firstly, we use an open-access OBIs dataset provided by the Key Laboratory of the Ministry of Education for Oracle Information Processing, Anyang Normal University~\cite{OBIsdetect2}. It has collected 9500 OBI rubbings (up to now) by a high-resolution scanner and then labeled every character with the upper left and lower right coordinates. Notably, this work only focuses on the detection task, so the number of classes in this dataset is all regarded as one.

Secondly, we use OBIMD alse provided by the Key Laboratory of the Ministry of Education for Oracle Information Processing, Anyang Normal University~\cite{obidataset2}. It is the world’s first comprehensive dataset encompassing multi-view information for OBI research, which has collected 10077 OBI rubbings. 
They are our original input datasets, which are used to train the model for the OBIs detection task.In the following sections, we will refer to the former and the latter as OBIs detection dataset and OBIMD respectively.

\textbf{OBC font library Dataset.} Besides, we use another dataset also provided by the Key Laboratory of the Ministry of Education for Oracle Information Processing, Anyang Normal University. It is OBC font library dataset called AYJGW~\cite{zitiku}, which is used as prior knowledge to enhance feature extraction. With the 
efforts of authoritative OBI experts, the glyph of every character in AYJGW is confirmed by its 
inherent meaning. Besides, the font models in the font library are all written by the calligraphy expert. 
In AYJGW, there are 3,881 character images which have uncontroversial font shapes, and it doesn't contain any noise. The dataset contains only images but not the annotation information of images, but our work requires the annotation information of its bounding box. Since our task is to detect the OBIs, and each image in this dataset is an independent OBC, the bounding box information of the image can be regarded as the bounding box annotation information of its object. The annotation information of the bounding box for each image in the dataset is formed by the above method.
\subsection{Overview}\label{subsec2}

\begin{figure}
\centering
\includegraphics[width=1.0\textwidth]{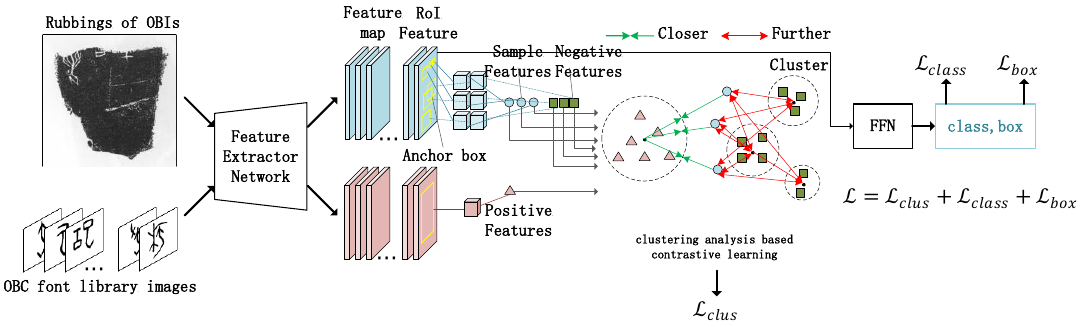}      
\caption{The pipeline of our proposed method. Our method takes an OBIs detection dataset and the OBC font library dataset as the inputs. The sample features and negative features are the features extracted in the positive and negative sample boxes on the OBIs images. The positive features are the features extracted in the positive sample boxes on the OBC font library dataset. 
}
\label{fig:reidd} 
\end{figure}

{The overall pipeline of our method is shown in Fig. ~\ref{fig:reidd}.}
Firstly, the OBIs dataset images and the OBC font library images are sent as inputs to the feature extraction network in the detection network at the same time. We will obtain feature maps after the forward propagation of the feature extraction network. We will get anchor boxes based on the feature map respectively, and extract the features of the mapped region of the anchor boxes on the feature map. After the above operations, we will obtain sample features and negative features from the OBIs dataset images, and positive features from the OBC font library images. These features are then flattened into feature vectors, and we will perform contrastive learning on these feature vectors. In contrastive learning, {we will calculate} a loss based on the sample and its positive and negative samples, and add this loss to the total loss of the network during training to adjust the model parameters by gradient feedback.

\subsection{Acquisition of RoI Features
}\label{subsec2}
Object detection networks usually first pass the image through a convolutional neural network to extract its feature map. For example, we have applied our method on Faster R-CNN~\cite{fasterrcnn}, DETR~\cite{DETR}, and Sparse R-CNN~\cite{sparsercnn}, which all use ResNet\cite{resnet} as their feature extraction network. First, the OBIs dataset images and the OBC font library images are sent as inputs to the feature extraction network in the detection network at the same time. Let's assume that the image from the OBIs dataset is $I_{1} \in \mathbb{R}^{H_{1} \times W_{1} \times 3 }$, the image from the OBC font library dataset is $I_{2} \in \mathbb{R}^{H_{2} \times W_{2} \times 3}$. We will obtain feature map $F_1=E(I_1) \in \mathbb{R}^{H_{1} \times W_{1} \times C_{1} }$ and feature map $F_2=E(I_2) \in \mathbb{R}^{H_{2} \times W_{2} \times C_{2} }$ respectively after the forward propagation of the feature extraction network. We will get anchor boxes based on the feature map, and we will set an {Intersection over union(IoU)} threshold to judge whether it is a positive anchor box or a negative anchor box according to whether the IoU values of the anchor box and ground truth bounding box are larger or smaller than it. The setting of the threshold is related to whether the original detection model contains sparse box property. If the original detection model contains sparse box property, the threshold will be set relatively small. On the contrary, if the original model does not contain sparse box property, that is, it contains dense box property, then the threshold will be set relatively large. We use the RoI Align layer to extract the features framed by coordinates on the feature map and scale them to a uniform size, and then flatten them to the feature vectors. At this time, the two groups of features to be clustered are the features of OBIs detection dataset images and the OBC font library images, and the next step is to perform cluster analysis.

\subsection{Clustering-based Contrastive Learning}
\label{subsec2}
Our method uses the K-Means clustering algorithm in the training phase to cluster the feature vectors corresponding to the negative anchor boxes in the OBIs detection dataset images and the positive anchor boxes in the OBC font library images respectively. We assume that the number of clusters of the features of the negative anchor boxes in the OBIs dataset is set to be N, we will get cluster centers $\left \{C_{N}\right \}$ by using the following formula after K-Means clustering. 
\begin{equation}
C_{N}=\frac{1}{\left|M_{N}\right|} \sum_{x_{i} \in M_{N}} x_{i}
\end{equation}
where $M_{N}$ is the set of feature points belonging to the cluster $C_{N}$.

Similarly, we assume that the number of clusters of the features of the positive anchor boxes in the OBC font library dataset is set to be M, we will get cluster centers $\left \{C_{M}\right \}$ after K-Means clustering, {then we will take the average at these cluster centers to obtain an average point ${C_{M}}^{mean}$.}
We'll conduct a contrastive learning of point and centers of clustering based on the clustering results, in order to achieve intra-cluster compactness and inter-cluster separation. In our method, the samples are the features of the positive anchor boxes in the OBIs images, {and their positive samples is ${C_{M}}^{mean}$}, they will form positive sample pairs. Their negative samples are {the cluster centers $\left \{C_{N}\right \}$}. We do this by using the following formula such that each feature point is close to its own cluster center and far away from other cluster centers.
\begin{equation}
\mathcal{L}_{\mathrm{clus}}\left(p_{n}\right)=-\log \frac{\exp \left(\boldsymbol{p}_{n} \cdot \boldsymbol{C_{M}}^{mean} / \tau\right)}{\sum_{n} \exp \left(\boldsymbol{p}_{n} \cdot \boldsymbol{C}_{N} / \tau\right)},
\end{equation}
where $\tau>0$ is a scalar temperature parameter, {${C_{M}}^{mean}$ refers to the average point of positive sample cluster center of feature point ${p}_{n}$}, which is from cluster centers $\left \{C_{M}\right \}$. ${C}_{N}$ refers to the negative sample cluster centers of feature point ${p}_{n}$, which is from cluster centers $\left \{C_{N}\right \}$. The numerator is the similarity of the feature point and its positive sample cluster center, the denominator is the similarity of the feature point and all of its positive sample cluster centers. At training time, we update the parameters in the direction of the descending gradient of the $\mathcal{L}_{\mathrm{clus}}$, to decrease the distance between ${p}_{n}$ and ${C}^{\prime}$, while increasing the distance between ${p}_{n}$ and ${C}_{n}$. The $\mathcal{L}_{\mathrm{clus}}$ will guide the model to train a feature network that makes different features more separate. In the case of our particular task are increasing the distance between the features of OBIs and the features of noise, cracks, and compound graphs. Compared with high-level semantic supervision information, it provides more effective and direct supervision from the aspect of data structure.

\subsection{Loss function}\label{subsec2}
The input of our method has two {image datasets}, namely the OBIs detection dataset images and the OBC font library images. During the forward propagation of the network, OBIs images will generate $\mathcal{L}_{\text {class}}$ and $\mathcal{L}_{\text {box}}$. The total loss function can be expressed as a weighted 
sum of $\mathcal{L}_{\text {class}}$,  $\mathcal{L}_{\text {box}}$ and $\mathcal{L}_{\text {clus}}$:
\begin{equation}
\mathcal{L}=\lambda_{1} \mathcal{L}_{\text {clus}}+\lambda_{2} \mathcal{L}_{class}+\lambda_{3} \mathcal{L}_{box}
\end{equation}

The OBC font library images are treated as prior knowledge to be introduced to guide the learning process of the model, which should only be used when calculating the $\mathcal{L}_{\text {clus}}$, and other losses they generate are not included in the final total loss. The weight parameters $\lambda_{1}$, $\lambda_{2}$, $\lambda_{3}$ are set differently according to the characteristics of each model.

\section{Results}\label{sec5}
\subsection{Implementation Details and Evaluation Metrics}\label{subsec2}
We applied our method to three detection models: Faster R-CNN, DETR, and Sparse R-CNN, and they are all implemented based on Python and PyTorch. We trained these models on four NVIDIA RTX A6000. We will introduce the details of experimental {implementation} for each of the three models. 

Faster R-CNN and Sparse R-CNN both used ResNet-50 as feature extraction network, then used Feature Pyramid {Networks (FPN)}\cite{FPN} to achieve better feature map fusion. We initialized the parameters of ResNet-50 with weights pre-trained on the ImageNet dataset Torchvision officially provided, and the other parameters were initialized randomly. We employed the SGD optimizer for training Faster R-CNN\cite{SGD}, with a momentum parameter
set to 0.9 and a weight decay parameter
set to $1 \times 10^{-4}$. 

DETR used ResNet-101 as the feature extraction network. We initialized the parameters of our DETR model with weights pre-trained on the COCO 2017 dataset Facebook officially provided. DETR and Sparse R-CNN both employed the AdamW\cite{AdamW} optimizer for training, with a weight decay parameter
set to $1 \times 10^{-4}$. 

The remaining parameter {settings} of the three models are shown in the {table \ref{tab:addlabel0}}.

\begin{table}[htbp]
  \centering
  \caption{The setting of learning rate, $\lambda_{1}$ , $\lambda_{2}$ , $\lambda_{3}$ , and Batchsize of Faster R-CNN+ours, DETR+ours and Sparse R-CNN+ours. }
  \label{tab:addlabel0}%
    \begin{tabular}{p{13.465em}ccc}
    \toprule
    Model & Learning Rate & $\lambda_{1}$ , $\lambda_{2}$ , $\lambda_{3}$ & Batchsize   \\
    \midrule
    Ours (Faster R-CNN) & $1 \times 10^{-2}$ & 1, 1, 1 & 2         \\
    Ours (DETR)  & $1 \times 10^{-4}$ & 1, 5, 1 & 2         \\
    Ours (Sparse R-CNN) & $2.5 \times 10^{-5}$ & 0.1, 1, 1 & 4        \\
    \bottomrule
    \end{tabular}%
\end{table}%

Before introducing the evaluation metrics of the method, we first introduce the concept of confusion matrix. The sample is a very important concept in the evaluation of object detection methods. Positive sample is the object that needs to be detected, and negative sample is the object that does not need to be detected. When the model is predicting, there are usually four situations: TP, FP, TN, and FN. In object detection, the final bounding boxes predicted by the model are regarded as positive samples, while whether the prediction belongs to TP or FP depends on the value of {IoU}. IoU calculates the ratio between the intersection area of the predicted bounding box and the ground truth bounding box and the merging area, and we will set a threshold value, which is 0.5 in this study. If the value of IoU of the predicted bounding box and ground truth bounding box is greater than this threshold, the predicted case of this bounding box is TP. Conversely, if the value of IoU of the predicted bounding box and any ground truth bounding box is less than this threshold, then the prediction of this bounding box is FP. If the IoU value of a ground truth bounding box and any predicted bounding box is less than this threshold, it is FN. 

We use three evaluation metrics~\cite{precision} commonly used in object detection models based on confusion matrix, namely Precision, {defined as} is the proportion of TP in all prediction numbers;
Recall, {defined as} is the proportion of TP in all object numbers;
F1-score, {defined as} is the harmonic mean of precision and recall, which is based on the comprehensive evaluation model of precision and recall.

\subsection{Main results}\label{subsec2}

To prove the performance improvement of our method for detection models, we conducted a series of comparative experiments on the two OBIs detection datasets mentioned in Section \textcolor{blue}{2.1}, Precision, Recall, and F1-socre were used to evaluate and compare the performance of the Faster R-CNN, DETR, DENO\cite{DENO} and Sparse R-CNN model and the improved Faster R-CNN, DETR, DENO, and Sparse R-CNN model, as shown in Table \ref{tab:addlabel1} and Table \ref{tab:addlabel2}. The model without special marking means that its backbone is ResNet.

\vspace{0.5cm}
\noindent
\begin{minipage}{\textwidth}
    \centering
    \captionof{table}{{Comparison results on OBIs detection dataset~\cite{OBIsdetect2} between existing models and our proposed method. AP is averaged over multiple loU values. Specifically we use 10 loU thresholds of .50:.05:.95. AP$_{50}$ is computed the average precision at a single loU of 0.50. AR$_{50}$ is computed the average recall at a single loU of 0.50.}}
    \label{tab:addlabel1}
    
    \resizebox{\textwidth}{!}{%
    \begin{tabular}{p{13.465em}ccccc}
    \toprule
    Model & AP & AP$_{50}$ & AP$_{75}$ & AR$_{50}$ & F1-Score$_{50}$ \\
    \midrule
    Faster R-CNN-R50 & 47.0 & 83.7 & 47.9    & 89.1    & 86.3 \\
    Faster R-CNN-Swin & 48.0 & 86.8 & 48.8    & 61.1    & 71.7 \\
    DETR-R101  & 35.7 & 75.8 & 28.4    & 88.2    & 81.5 \\
    DENO-Swin  & 39.9 & 78.4 & 36.1    & 91.7    & 84.5 \\
    Sparse R-CNN-R50 & 36.5 & 72.3 & 33.1    & 89.0    & 79.7 \\
    Sparse R-CNN-Swin & 36.5 & 71.4 & 33.3    & 45.9    & 55.9 \\
    \textbf{Ours} (Faster R-CNN-R50) & 48.1\textcolor{blue}{(+1.1)} & 86.4\textcolor{blue}{(+2.7)} & 48.7\textcolor{blue}{(+0.8)}    & 92.5\textcolor{blue}{(+3.4)}    & 89.3\textcolor{blue}{(+3.0)} \\
    \textbf{Ours} (Faster R-CNN-Swin) & 47.8\textcolor{red}{(-0.2)} & 86.6\textcolor{red}{(-0.2)} & 48.2\textcolor{red}{(-0.6)}    & 62.4\textcolor{blue}{(+1.3)}    & 72.5\textcolor{blue}{(+0.8)} \\
    \textbf{Ours} (DETR-R101) & 36.7\textcolor{blue}{(+1.0)}    & 77.0\textcolor{blue}{(+1.2)}    & 30.7\textcolor{blue}{(+2.3)} & 88.3\textcolor{blue}{(+0.1)}    & 82.3\textcolor{blue}{(+0.8)} \\
    \textbf{Ours} (DENO-Swin) & 41.7\textcolor{blue}{(+1.8)}    & 81.3\textcolor{blue}{(+2.9)}    & 38.6\textcolor{blue}{(+2.5)} & 92.5\textcolor{blue}{(+0.8)}    & 86.5\textcolor{blue}{(+2.0)} \\
    \textbf{Ours} (Sparse R-CNN-R50) & 37.2\textcolor{blue}{(+0.7)} & 73.4\textcolor{blue}{(+1.1)} & 34.1\textcolor{blue}{(+1.0)}    & 89.0    & 80.4\textcolor{blue}{(+0.7)} \\
    \textbf{Ours} (Sparse R-CNN-Swin) & 37.2\textcolor{blue}{(+0.7)} & 72.4\textcolor{blue}{(+1.0)} & 34.6\textcolor{blue}{(+1.3)}    & 47.0\textcolor{blue}{(+1.1)}    & 57.0\textcolor{blue}{(+1.1)} \\
    \bottomrule
    \end{tabular}%
    }
\end{minipage}
\vspace{0.5cm}

\vspace{0.5cm}
\noindent
\begin{minipage}{\textwidth}
    \centering
    \captionof{table}{Comparison results on OBIMD~\cite{obidataset2} between existing models and our proposed method.}
    \label{tab:addlabel2}
    
    \resizebox{\textwidth}{!}{%
    \begin{tabular}{p{13.465em}ccccc}
    \toprule
    Model & AP & AP$_{50}$ & AP$_{75}$ & AR$_{50}$ & F1-Score$_{50}$ \\
    \midrule
    Faster R-CNN-R50 & 33.2 & 70.9 & 26.2    & 79.8    & 75.1  \\
    DETR-R101  & 29.3 & 66.4 & 21.0    & 75.7    & 70.7  \\
    Sparse R-CNN-R50 & 29.2 & 60.1 & 23.3    & 77.7    & 67.7  \\
    \textbf{Ours} (Faster R-CNN-R50) & 34.3\textcolor{blue}{(+1.1)} & 73.4\textcolor{blue}{(+2.5)} & 26.6\textcolor{blue}{(+0.4)}    & 81.4\textcolor{blue}{(+1.6)}    & 77.2\textcolor{blue}{(+2.1)} \\
    \textbf{Ours} (DETR-R101) & 30.0\textcolor{blue}{(+0.7)}    & 68.0\textcolor{blue}{(+1.6)}    & 22.0\textcolor{blue}{(+1.0)} & 76.1\textcolor{blue}{(+0.4)}    & 71.8\textcolor{blue}{(+1.1)} \\
    \textbf{Ours} (Sparse R-CNN-R50) & 30.1\textcolor{blue}{(+0.9)} & 62.6\textcolor{blue}{(+2.5)} & 24.7\textcolor{blue}{(+1.4)}    & 78.7\textcolor{blue}{(+1.0)}    & 69.7\textcolor{blue}{(+2.0)} \\
    \bottomrule
    \end{tabular}%
    }
\end{minipage}
\vspace{0.5cm}

\begin{table}[htbp]
  \centering
  \caption{Comparison Training time and Memory usage on OBIs detection dataset~\cite{OBIsdetect2} and OBIMD~\cite{obidataset2} between existing models and our proposed method.}
  \label{tab:addlabel3}%
    \begin{tabular}{p{13.465em}ccc}
    \toprule
    Dataset & Model & Training time & Memory usage(GiB)
    \\
    \midrule
    OBIs detection dataset & Faster R-CNN & 10h56min & 18.77          \\
    & \textbf{Ours} (Faster R-CNN)  & 13h23min & 20.57          \\
    & DETR & 13h12min & 10.50        \\
    & \textbf{Ours} (DETR)  & 17h33min & 13.01          \\
    & Sparse R-CNN & 7h32min & 12.31         \\
    & \textbf{Ours} (Sparse R-CNN) & 9h42min & 16.14         \\
    \midrule
    OBIMD & Faster R-CNN & 14h1min & 18.89          \\
    & \textbf{Ours} (Faster R-CNN)  & 16h27min & 20.90          \\
    & DETR & 15h8min & 10.62         \\
    & \textbf{Ours} (DETR)  & 20h25min & 13.23          \\
    & Sparse R-CNN & 9h35min & 12.50         \\
    & \textbf{Ours} (Sparse R-CNN) & 11h45min & 16.24         \\
    \bottomrule
    \end{tabular}%
\end{table}%

{As shown in Table \ref{tab:addlabel1} and Table \ref{tab:addlabel2}, Considering transformer based approaches like Swin Transformer can provide deeper insights into the advantages of clustering based representation learning, we added the experiment of replacing the backbone of the three models with Swin Transformer~\cite{swin}. Among them, because DETR replaced the backbone with the Swin Transformer and there was no global pre-training weight, the model failed to converge during training. So we replaced DETR with DINO: DETR with Improved DeNoising Anchor Boxes for End-to-End Object Detection~\cite{DENO}. It is an improved version of DETR and can converge better during training. 
From the numerical point of view, after the application of our proposed method in these three models, the precision, recall, and F1-score have been improved to some extent compared with the baseline model, only the accuracy rate on Faster R-CNN-Swin has slightly decreased. so our proposed method is meaningful for the improvement of the detection effect of the models.} 

{As shown in Table \ref{tab:addlabel3}, from the numerical point of view, after the application of our proposed method in these three models, the consumption in terms of both training time and memory usage have increased to a certain extent, but it is all within an acceptable range.}

{As is well known, there are some high-performance and most popular detection
frameworks such as YOLO~\cite{YOLO}. However, our method is not suitable for application to models of the YOLO class. Because YOLO ultimately uses global features to predict categories and regression box coordinates, while the method we use takes out each feature in the box to predict categories and regression box coordinates, and our clustering algorithm improves for each feature, it has no effect on global features.}

In addition to the comparison of numerical results, we also did visualization experiments, namely the visualization of features and the visualization of detection results.

\textbf{Detection results.} We visualized the detection results of the three models. As shown in Fig. \ref{fig:three_images} and Fig. \ref{fig:three_images2}, it can be obviously seen that our method has significantly improved the detection effect of the three models.

\setlength{\figheight}{(\textheight - 1cm) / 3}  

\begin{figure}[p]  
    \centering
    \begin{subfigure}[b]{\textwidth}
        \centering
        \includegraphics[height=\figheight]{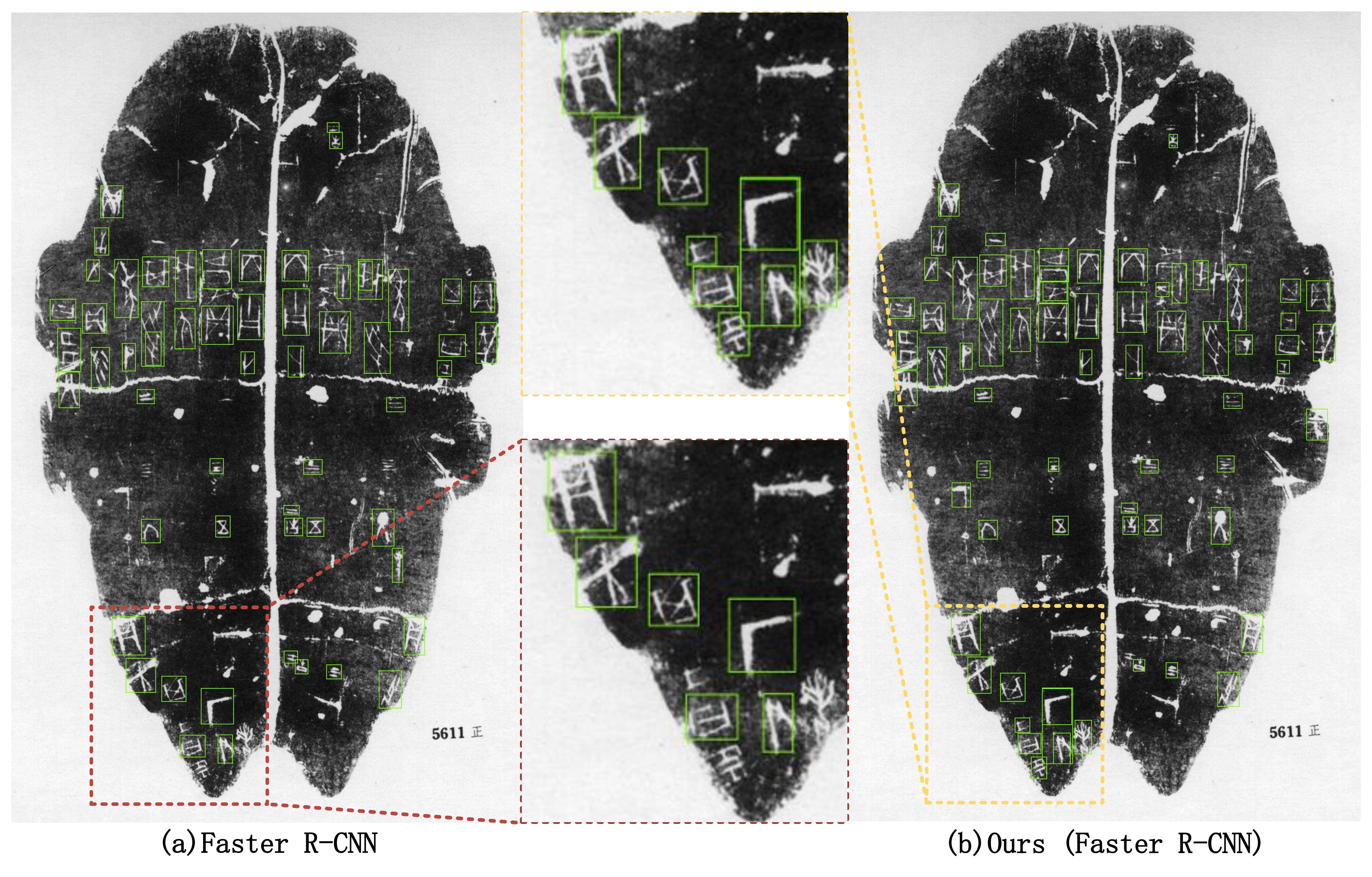}      
        \caption{Comparative detection results on Faster R-CNN}
        \label{fig:intro1}
    \end{subfigure}
    
    \vspace{0.1cm}  
    
    \begin{subfigure}[b]{\textwidth}
        \centering
        \includegraphics[height=\figheight]{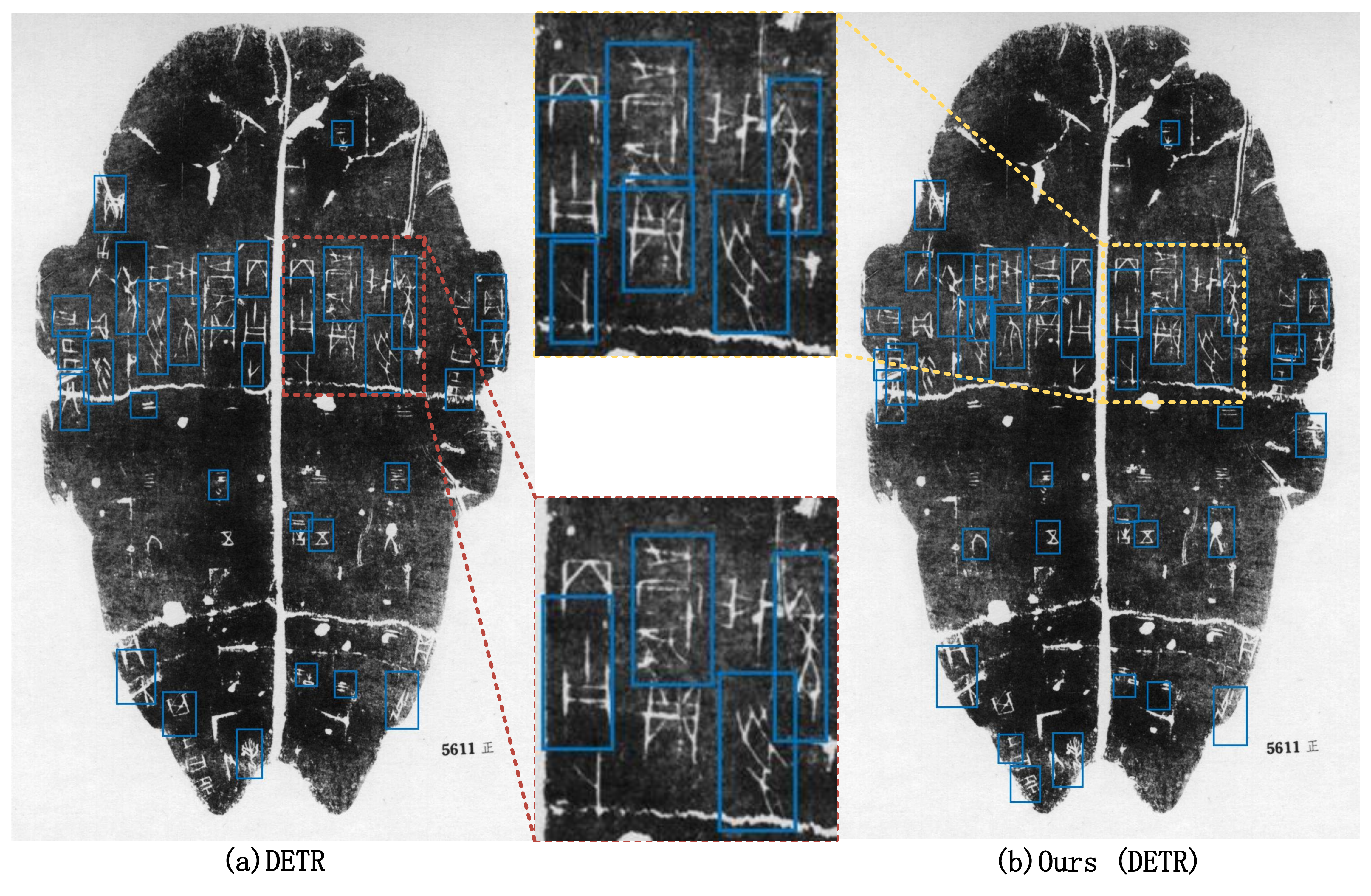}      
        \caption{Comparative detection results on DETR}
        \label{fig:intro2}
    \end{subfigure}
    
    \vspace{0.1cm}  
    
    \begin{subfigure}[b]{\textwidth}
        \centering
        \includegraphics[height=\figheight]{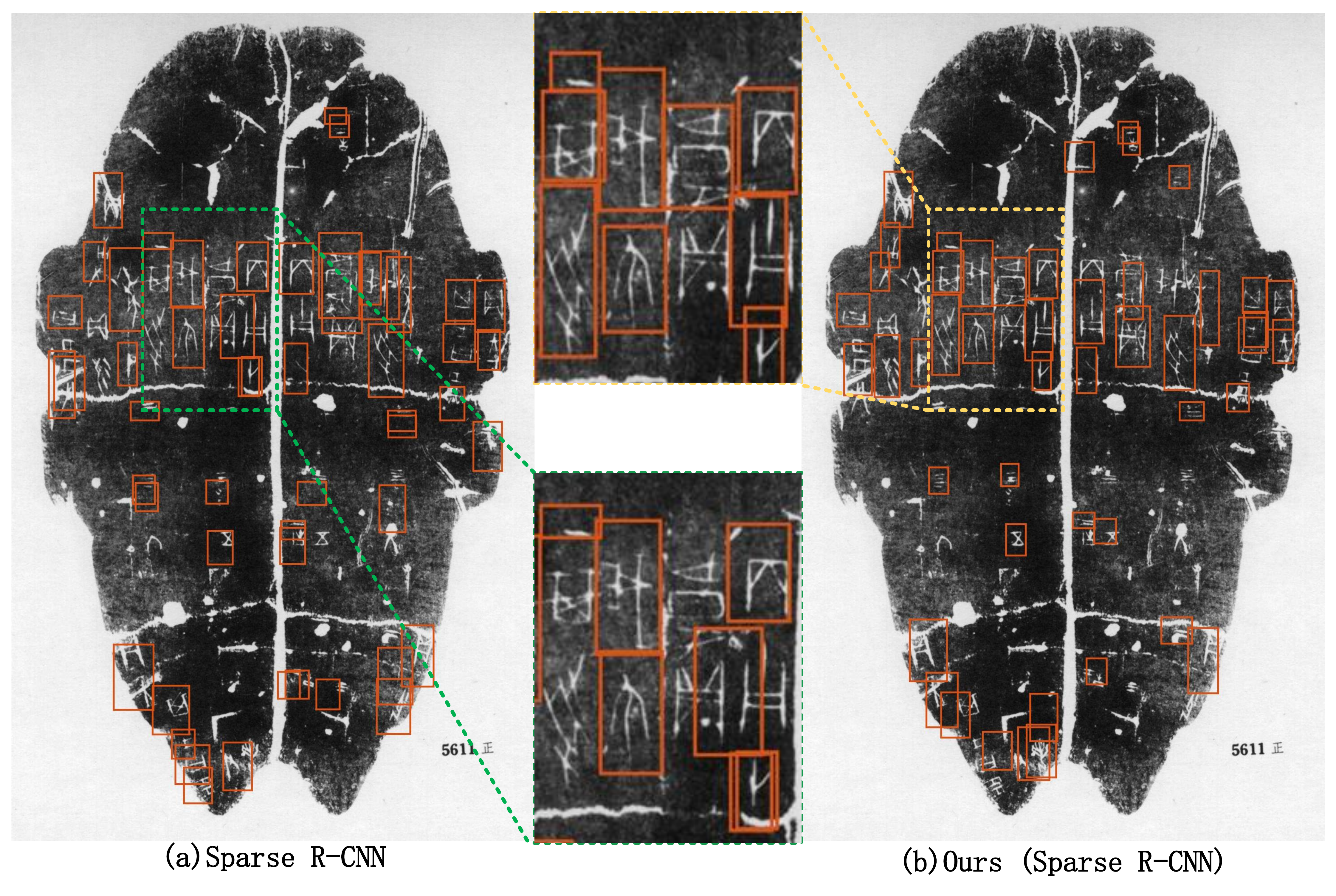}      
        \caption{Comparative detection results on Sparse R-CNN}
        \label{fig:intro3}
    \end{subfigure}
    
    \caption{{Comparative detection results on three models, the partially enlarged image at the bottom is the baseline model, and the one at the top is with our method.}}
    \label{fig:three_images}
\end{figure}

\setlength{\figheight}{(\textheight - 1cm) / 3}  

\begin{figure}[p]  
    \centering
    \begin{subfigure}[b]{\textwidth}
        \centering
        \includegraphics[height=\figheight]{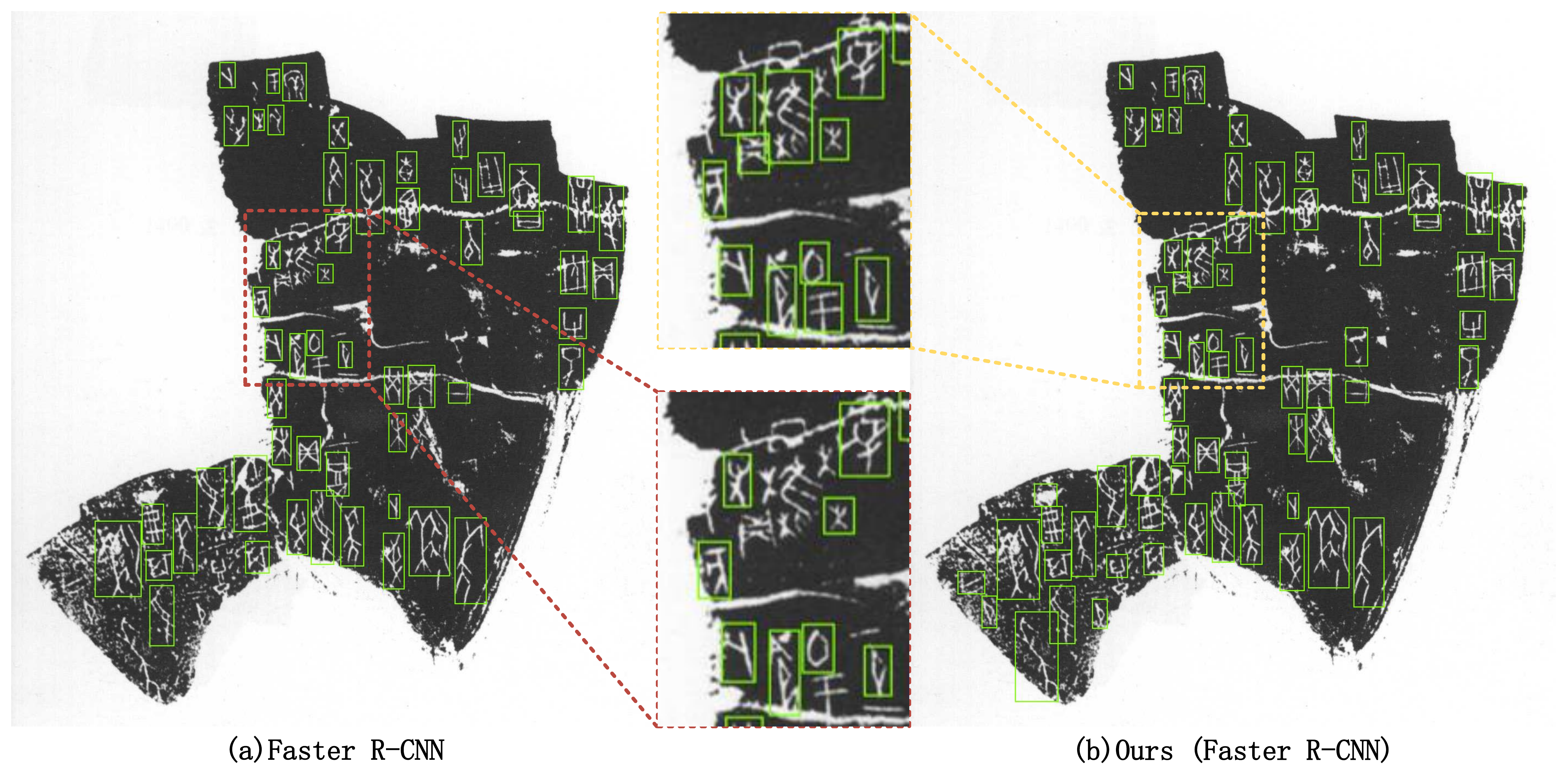}      
        \caption{Comparative detection results on Faster R-CNN}
        \label{fig:intro1}
    \end{subfigure}
    
    \vspace{0.1cm}  
    
    \begin{subfigure}[b]{\textwidth}
        \centering
        \includegraphics[height=\figheight]{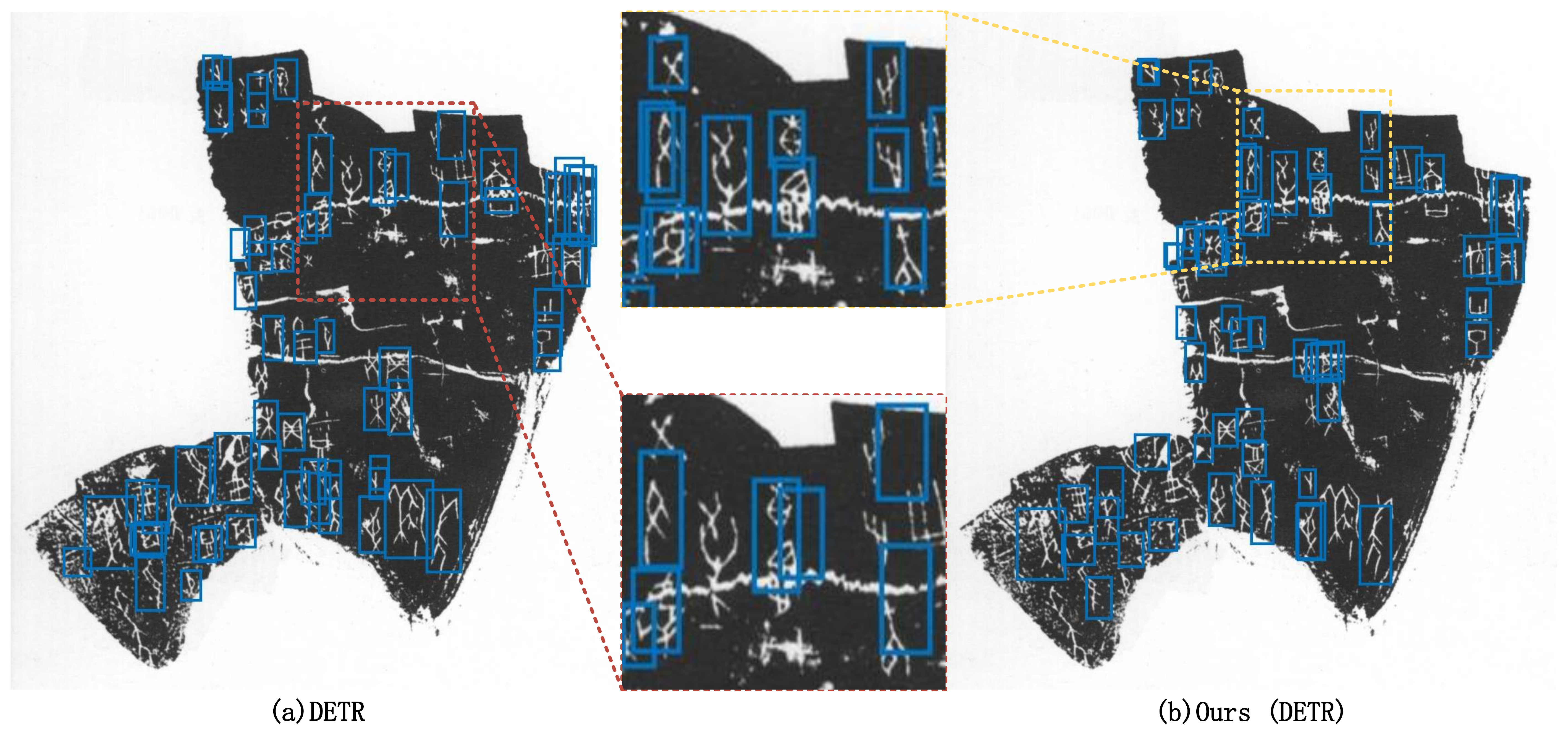}      
        \caption{Comparative detection results on DETR}
        \label{fig:intro2}
    \end{subfigure}
    
    \vspace{0.1cm}  
    
    \begin{subfigure}[b]{\textwidth}
        \centering
        \includegraphics[height=\figheight]{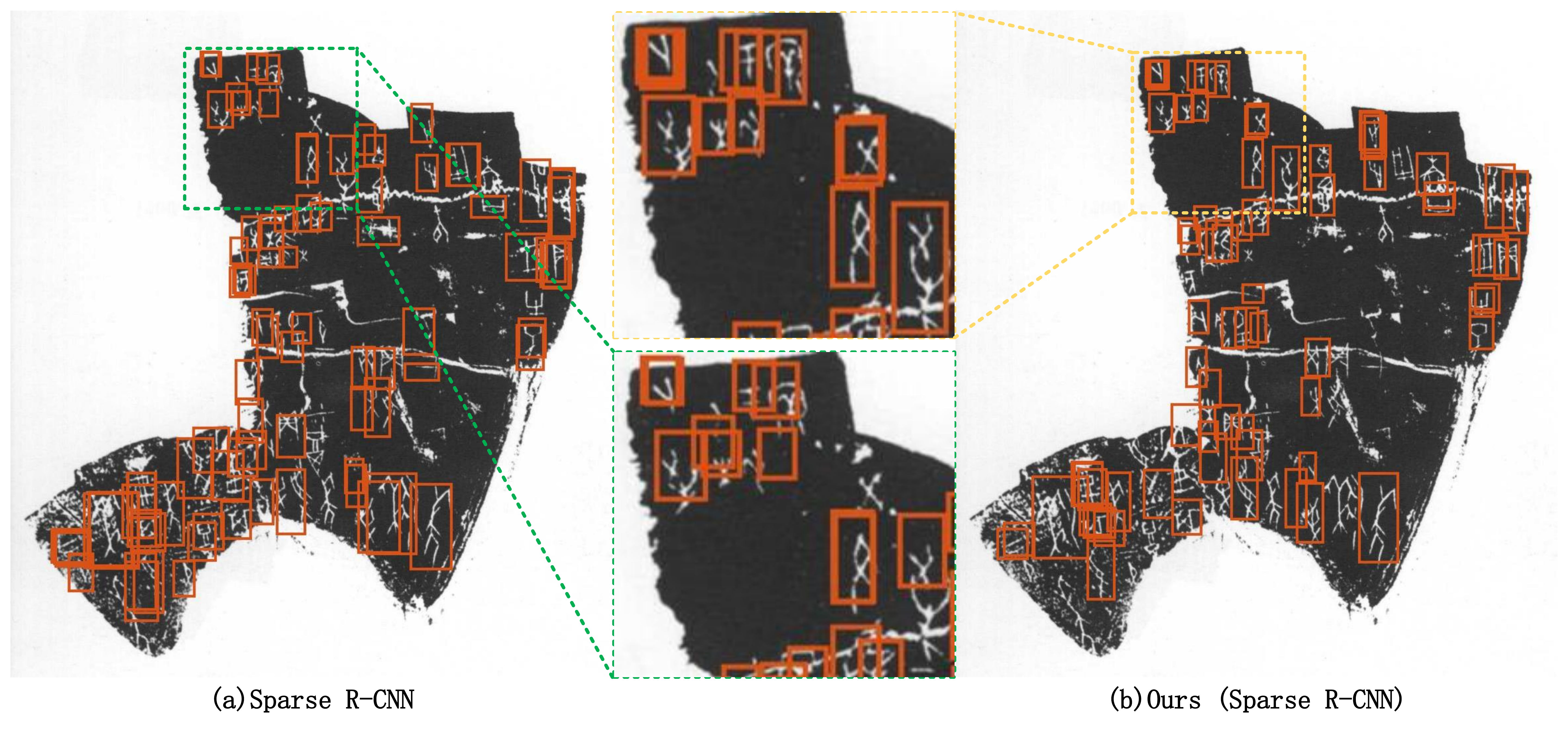}      
        \caption{Comparative detection results on Sparse R-CNN}
        \label{fig:intro3}
    \end{subfigure}
    
    \caption{{Comparative detection results on three models, the partially enlarged image at the bottom is the baseline model, and the one at the top is with our method.}}
    \label{fig:three_images2}
\end{figure}

\textbf{Features.} The feature vectors for comparison learning were sample features, positive features, and negative features, which were reduced to two dimensions using t-SNE~\cite{tsne} dimensionality reduction method, and these points were presented in an image. We can use these visualizations to further visualize the role of contrast learning in our method.

\begin{figure}
\centering
\includegraphics[width=1.0\textwidth]{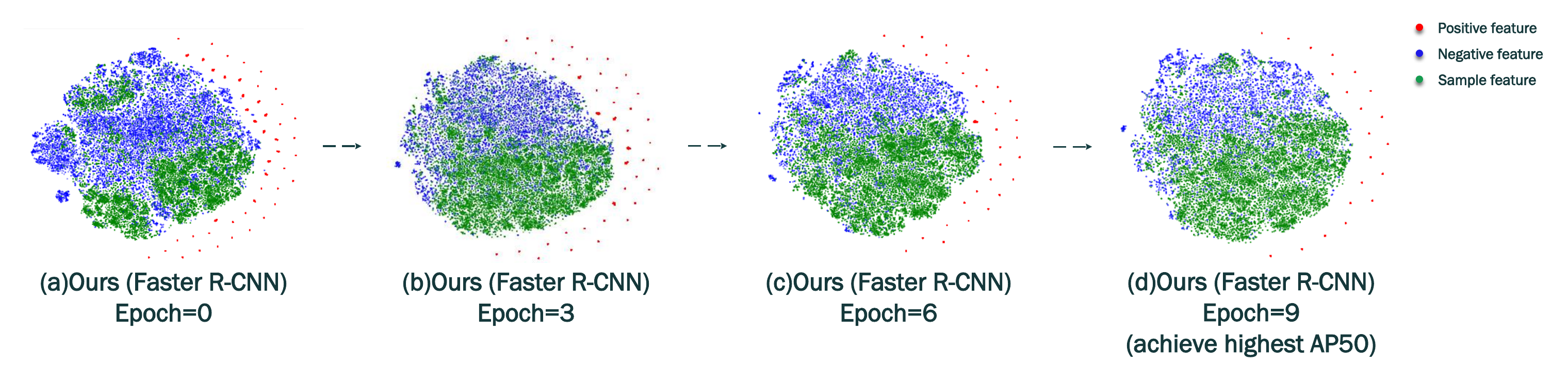}      
\caption{Comparison of visualization results of the distribution of positive features, negative features and sample features generated from the Faster R-CNN+ours model which uses weight from epoch=0, 3, 6, 9.}
\label{figepoch} 
\end{figure}

\begin{figure}
\centering
\includegraphics[width=0.6\textwidth, height=8cm, keepaspectratio]{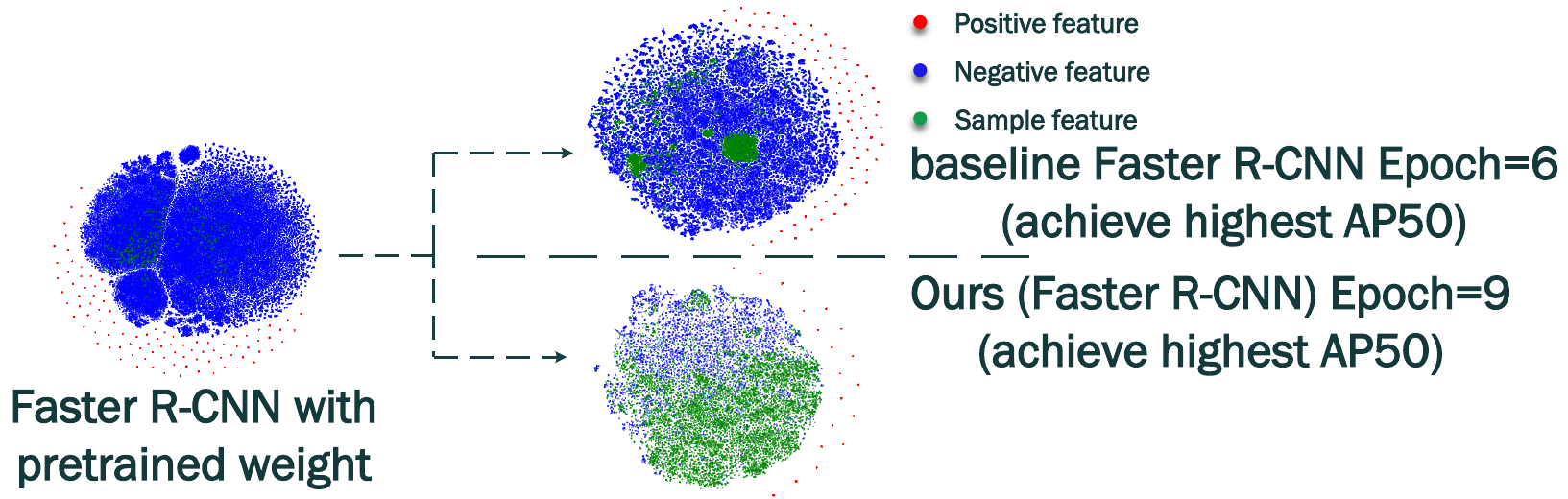}      
\caption{Comparison of visualization results of the distribution of positive features, negative features and sample features generated from the baseline Faster R-CNN model which uses weight that reaches the highest AP$_{50}$ and positive features, negative features and sample features generated from the Faster R-CNN+ours model which uses weight that reaches the highest AP$_{50}$.}.
\label{fig:reid4} 
\end{figure}

As shown in Fig. \ref{figepoch}, we observe the visualization of the evolution process of the Faster R-CNN model using our proposed method first. With the epoch number increased, the sample features, which were initially scattered and unordered, mixed with negative features, are gradually guided by positive features to concentrate at the bottom and linearly separate from negative features. Throughout the process, the number of sample features increases gradually, while the number of negative features decreases. The visualization results of this evolution indicate that the features from OBC font library dataset are effective in guiding the features from OBIs dataset to be away from those of non-characters. Furthermore, as shown in {Fig.} \ref{fig:reid4}, by directly comparing the feature point visualization image of the baseline Faster R-CNN model and the Faster R-CNN model using our proposed method, it can be seen that the sample features and negative features in the baseline Faster R-CNN model are mixed together and linearly inseparable. In contrast, the sample features and negative features in the Faster R-CNN model using our method are separated and linearly separable. Moreover, the number of sample features in the latter is much greater than that in the former, and the number of negative features in the latter is much smaller than that in the former, which indicates that applying our method can guide the model to output more anchor boxes containing characters, while reducing the output of anchor boxes containing cracks, noise, etc. This has a certain significance for improving the detection performance of the model.

\subsection{Hyperparameter and Result analysis}\label{subsec2}
Firstly, we analyze the different effects of the setting of four hyper-parameters, namely the number of OBC from OBC font library dataset used as prior knowledge, the setting of temperature parameter $\tau$, the setting of $\lambda_{1}$ , $\lambda_{2}$ , $\lambda_{3}$, and the setting of clustering technique on the model detection effect. Secondly, we analyze whether the results are reproducible.

\textbf{Number of OBC used as prior knowledge.}
We applied our method on the Faster R-CNN, DETR and Sparse R-CNN models with the number of OBC used as prior knowledge sets of 10, 20, and 50 respectively, and the results obtained {are shown in Table \ref{OBC0}}.

Firstly, we present the experimental results of different numbers of OBC used as prior knowledge on the Faster R-CNN model. When the number of OBC used as prior knowledge is set to 10 and 20, the performance on AP, AP$_{50}$, and AP$_{75}$ is improved compared to the baseline Faster R-CNN model. The improvement effect of the number of OBC used as prior knowledge set to 20 is better than that of the number of OBC used as prior knowledge set to 10. When the number of OBC used as prior knowledge is set to 50, the performance on AP and AP$_{50}$ is slightly decreased compared to the baseline Faster R-CNN model, and the performance on AP$_{75}$ is basically the same. Therefore, after the number of OBC used as prior knowledge is set to 50, our method has a negative effect on the Faster R-CNN model, resulting in lower accuracy than the baseline Faster R-CNN model. 

\begin{figure}[ht]
    \centering
    \begin{subfigure}[b]{0.32\textwidth}
        \includegraphics[width=\textwidth]{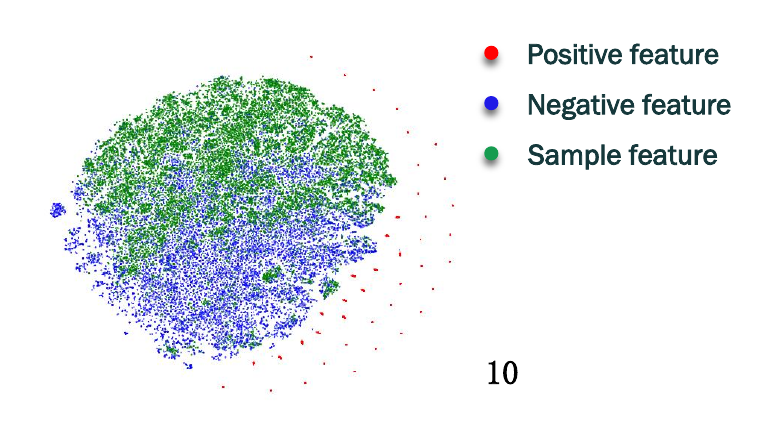}
        \caption{}
        \label{fig:sub1}
    \end{subfigure}
    \hfill
    \begin{subfigure}[b]{0.32\textwidth}
        \includegraphics[width=\textwidth]{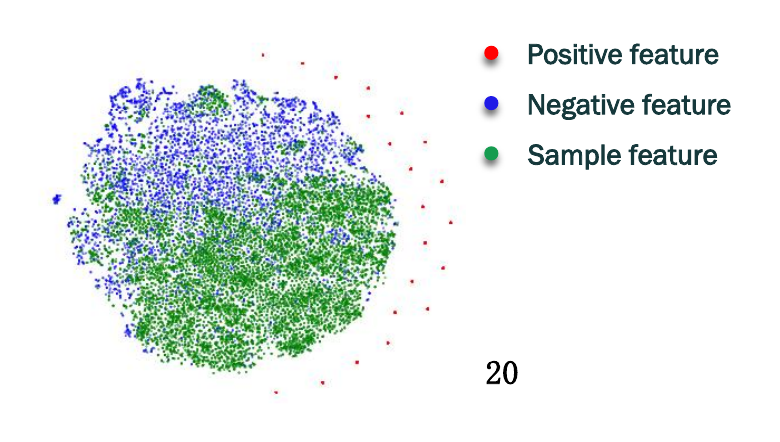}
        \caption{}
        \label{fig:sub2}
    \end{subfigure}
    \hfill
    \begin{subfigure}[b]{0.32\textwidth}
        \includegraphics[width=\textwidth]{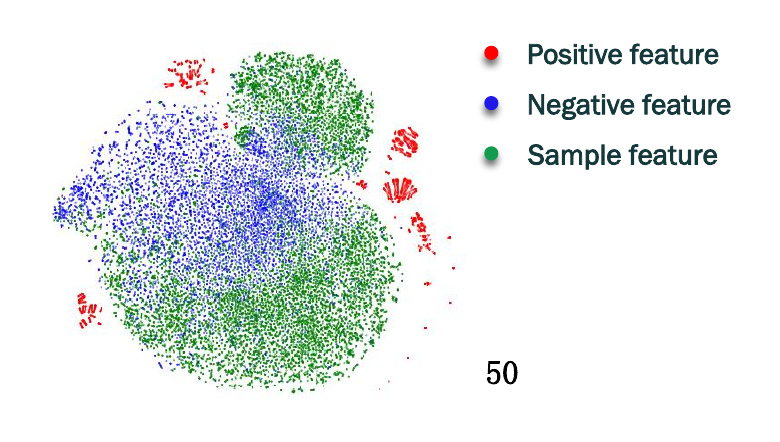}
        \caption{}
        \label{fig:sub3}
    \end{subfigure}
    \caption{Comparison of visualization results of the distribution of positive features, negative features and sample features generated from the Faster R-CNN+ours model with the number of OBC used as prior knowledge is set to 10, 20, 50.}
    \label{fig:overall}
\end{figure}

We perform a visual analysis of this experimental result by visualizing the distribution of sample features, positive features, and negative features when the number of OBC used as prior knowledge is set to 10, 20, and 50. As shown in Fig. \ref{fig:overall}, when the number of OBC used as prior knowledge is set to 10 and 20, the positive features are concentrated on the right side, providing effective guidance for the distribution of sample features, and finally, sample features and negative features can be linearly separated. When the number of OBC used as prior knowledge is set to 50, the positive features are distributed relatively randomly, providing negative guidance for the distribution of sample features, and finally, sample features and negative features cannot be linearly separated.

Secondly, we look at the experimental results of different number of OBC used as prior knowledge on the DETR model. When the number of OBC used as prior knowledge is set to 10, there is a decrease in the performance compared with baseline DETR model on AP, AP$_{50}$ and AP$_{75}$. When the number of OBC used as prior knowledge is set to 20 and 50, the effect on AP, AP$_{50}$ and AP$_{75}$ is improved compared with baseline DETR model. In terms of overall improvement effect, {the number of OBC used} as prior knowledge set to 50 is better than the set to 20.

The experimental results of different numbers of OBC used as prior knowledge on the Sparse R-CNN model are similar to that for the DETR model. The best setting is 20.

\vspace{0.5cm}
\noindent
\begin{minipage}{\textwidth}
    \centering
    \captionof{table}{Effect of number of OBC used as prior knowledge on different models. The performance of the number of OBC used as prior knowledge set to 0 means the baseline performance.}
    \label{OBC0}
    
    \resizebox{\textwidth}{!}{%
    \begin{tabular}{c cccc ccc ccc}
        \toprule
        \multirow{2}{*}{OBC} & \multicolumn{3}{c}{Faster R-CNN} & \multicolumn{3}{c}{DETR} & \multicolumn{3}{c}{Sparse R-CNN} \\
        \cmidrule(lr){2-4} \cmidrule(lr){5-7} \cmidrule(lr){8-10}
         & AP & AP$_{50}$ & AP$_{75}$ & AP & AP$_{50}$ & AP$_{75}$ & AP & AP$_{50}$ & AP$_{75}$ \\
        \midrule
        0   & 47.0 & 83.7 & 47.9 & 35.7 & 75.8 & 28.4 & 36.5 & 72.3 & 33.1 \\
        10  & 47.5\textcolor{blue}{(+0.5)} & 85.2\textcolor{blue}{(+1.5)} & 49.1\textcolor{blue}{(+1.2)} 
            & 35.2\textcolor{red}{(-0.5)} & 75.5\textcolor{red}{(-0.3)} & 27.8\textcolor{red}{(-0.6)}
            & 36.3\textcolor{red}{(-0.2)} & 71.7\textcolor{red}{(-0.6)} & 33.1 \\
        20  & 48.1\textcolor{blue}{(+1.1)} & 86.4\textcolor{blue}{(+2.7)} & 48.7\textcolor{blue}{(+0.8)}
            & 36.0\textcolor{blue}{(+0.3)} & 76.7\textcolor{blue}{(+0.9)} & 28.9\textcolor{blue}{(+0.5)}
            & 37.2\textcolor{blue}{(+0.7)} & 73.4\textcolor{blue}{(+1.1)} & 34.1\textcolor{blue}{(+1.0)} \\
        50  & 46.7\textcolor{red}{(-0.3)} & 83.2\textcolor{red}{(-0.5)} & 48.0\textcolor{blue}{(+0.1)}
            & 36.8\textcolor{blue}{(+1.1)} & 76.5\textcolor{blue}{(+0.7)} & 31.1\textcolor{blue}{(+2.7)}
            & 36.7\textcolor{blue}{(+0.2)} & 72.4\textcolor{blue}{(+0.1)} & 33.4\textcolor{blue}{(+0.3)} \\
        \bottomrule
    \end{tabular}%
    }
\end{minipage}
\vspace{0.5cm}

\textbf{Setting of temperature parameter $\boldsymbol{\tau}$.} Our method was experimented on three models and proved to improve the detection performance of all three models compared to the original baseline model. However, it can be observed that the improvement in the sparse detection method is not as significant as that in the dense detection method. The reason for this result is that in the contrastive learning process of the dense detection method in our method, each sample has more negative samples available for learning, while the number of negative samples corresponding to each sample in the sparse detection method is relatively smaller. In contrastive learning, the temperature plays a role in controlling the strength of penalties on the hard negative samples. Specifically, contrastive loss with small temperature tends to penalize much more on the hardest negative samples. On the other hand, contrastive loss with large temperatures is less sensitive to the hard negative samples. In our method, there is a problem that the number of negative samples corresponding to samples is small in the sparse detection model, so the distinction between positive and negative samples should be enhanced, and the smaller the corresponding temperature should be set, the better the experimental results will be. We did the following experiment.

\begin{table}[htbp]
  \centering
  \caption{Effect of Setting of temperature parameter $\boldsymbol{\tau}$ on different models. The performance of temperature parameter $\boldsymbol{\tau}$ set to 0.1 is viewed as the baseline performance.}
  \label{ttable}
  \setlength{\tabcolsep}{4pt}
  \begin{tabular}{c c cccc ccc}
    \toprule
    \multirow{2}{*}{OBC} & \multirow{2}{*}{$\boldsymbol{\tau}$}& \multicolumn{3}{c}{Ours (DETR)} & \multicolumn{3}{c}{Ours (Sparse R-CNN)} \\
    \cmidrule(lr){3-5} \cmidrule(lr){6-8} 
     & & AP & AP$_{50}$ & AP$_{75}$ & AP & AP$_{50}$ & AP$_{75}$  \\
    \midrule
    10  &0.1& 35.2 & 75.5 & 27.8 
        & 36.3 & 71.7 & 33.1
        \\
    10  &0.05& 36.7\textcolor{blue}{(+1.5)} & 77.0\textcolor{blue}{(+1.5)} & 30.7\textcolor{blue}{(+2.9)} 
        & 37.3\textcolor{blue}{(+1.0)} & 72.8\textcolor{blue}{(+1.1)} & 34.7\textcolor{blue}{(+1.6)}
        \\
    10  &0.01& 36.5\textcolor{blue}{(+1.3)} & 76.1\textcolor{blue}{(+0.6)} & 30.8\textcolor{blue}{(+3.0)} 
        & 36.5\textcolor{blue}{(+0.2)} & 72.0\textcolor{blue}{(+0.3)} & 33.4\textcolor{blue}{(+0.3)}
        \\
    10  &0.005& 34.6\textcolor{red}{(-0.6)} & 75.1\textcolor{red}{(-0.4)} & 27.7\textcolor{red}{(-0.1)} 
        & 36.6\textcolor{blue}{(+0.3)} & 72.2\textcolor{blue}{(+0.5)} & 33.8\textcolor{blue}{(+0.7)}
        \\
    20  &0.1& 36.0 & 76.7 & 28.9 
        & 37.2 & 73.4 & 34.1
        \\
    20  &0.05& 37.4\textcolor{blue}{(+1.4)} & 76.8\textcolor{blue}{(+0.2)} & 32.4\textcolor{blue}{(+3.5)}
        & 36.9\textcolor{red}{(-0.3)} & 72.7\textcolor{red}{(-0.7)} & 33.1\textcolor{red}{(-1.0)}
        \\
    20  &0.01& 36.0 & 74.9\textcolor{red}{(-1.8)} & 30.3\textcolor{blue}{(+1.4)}
        & 37.0\textcolor{red}{(-0.2)} & 72.9\textcolor{red}{(-0.5)} & 33.4\textcolor{red}{(-0.7)}
        \\
    20  &0.005& 35.2\textcolor{red}{(-0.8)} & 75.2\textcolor{red}{(-1.5)} & 28.9
        & 36.6\textcolor{red}{(-0.6)} & 72.1\textcolor{red}{(-1.3)} & 33.4\textcolor{red}{(-0.7)}
        \\
    50  &0.1& 36.8 & 76.5 & 31.1
        & 36.7 & 72.4 & 33.4
        \\
    50  &0.05& 35.3\textcolor{red}{(-1.5)} & 75.8\textcolor{red}{(-0.7)} & 27.5\textcolor{red}{(-3.6)}
        & 35.9\textcolor{red}{(-0.8)} & 71.4\textcolor{red}{(-1.0)} & 32.5\textcolor{red}{(-0.9)}
        \\
    50  &0.01& 38.5\textcolor{blue}{(+1.7)} & 76.8\textcolor{blue}{(+0.3)} & 33.7\textcolor{blue}{(+2.6)}
        & 37.3\textcolor{blue}{(+0.6)} & 73.0\textcolor{blue}{(+0.6)} & 34.5\textcolor{blue}{(+1.1)}
        \\    
    50  &0.005& 34.3\textcolor{red}{(-2.5)} & 75.3\textcolor{red}{(-1.2)} & 26.2\textcolor{red}{(-4.9)}
        & 36.3\textcolor{red}{(-0.4)} & 71.7\textcolor{red}{(-0.7)} & 33.2\textcolor{red}{(-0.2)}
        \\    
    \bottomrule
  \end{tabular}%
\end{table}%


From Table \ref{ttable}, it can be seen that when the number of OBC used as prior knowledge is set to 10, the temperature parameter $\boldsymbol{\tau}$ decreased from 0.1 to 0.05 and 0.01, the performance on AP, AP$_{50}$, and AP$_{75}$ is improved compared to the baseline DETR and Sparse R-CNN model. The best setting of the temperature parameter $\boldsymbol{\tau}$ for the DETR and Sparse R-CNN model are both 0.05 when the number of OBC used as prior knowledge is set to 10.
{When the number of OBC used as prior knowledge is set to 20}, the temperature parameter $\boldsymbol{\tau}$ decreased from 0.1 to 0.05, the performance on AP, AP$_{50}$, and AP$_{75}$ is improved compared to the baseline DETR model. The best setting of the temperature parameter $\boldsymbol{\tau}$ for the DETR and Sparse R-CNN model are respectively 0.05 and 0.1 when the number of OBC used as prior knowledge is set to 20.
When the number of OBC used as prior knowledge is set to 50, the temperature parameter $\boldsymbol{\tau}$ decreased from 0.1 to 0.01, the performance on AP, AP$_{50}$, and AP$_{75}$ is improved {compared to the baseline DETR and Sparse R-CNN models}. The best setting of the temperature parameter $\boldsymbol{\tau}$ for the DETR and Sparse R-CNN model are both 0.01 when the number of OBC used as prior knowledge is set to 50.
Therefore, when our method is applied to the sparse detection model, the temperature should be set smaller to compensate for the problem that when the sparse detection models use our method, they will have fewer negative samples corresponding to the samples in the contrastive learning process. This will enable our method to achieve better performance when applied to the sparse detection model.

{To further optimize our method performance, we use Bayesian Optimization technique to tune our hyperparameters, and the results obtained were shown in Table \ref{BYS}.}

\vspace{0.5cm}
\noindent
\begin{minipage}{\textwidth}
    \centering
    \captionof{table}{Use Bayesian Optimization technique to tune our hyperparameters. The M and N are the number of clusters of negative samples and the number of clusters of positive samples mentioned in Section \textcolor{blue}{2.4}. }
    \label{BYS}
    
    \resizebox{\textwidth}{!}{%
    \begin{tabular}{p{13.465em}cccccccc}
    \toprule
    Model & OBC & $\boldsymbol{\tau}$ & M,N & AP & AP$_{50}$ & AP$_{75}$ & AR$_{50}$ & F1-Score$_{50}$ \\
    \midrule
    Faster R-CNN &  & & & 47.0 & 83.7    & 47.9    & 89.1 & 86.3   \\
    DETR  &  & &  & 35.7 & 75.8 & 28.4 & 88.2    & 81.5     \\
    Sparse R-CNN &  & &  & 36.5 & 72.3 & 33.1    & 89.0    & 79.7  \\
    \textbf{Ours} (Faster R-CNN) & 27 & 0.099 & 63,3 & 48.5\textcolor{blue}{(+1.5)} & 86.6\textcolor{blue}{(+2.9)} & 48.7\textcolor{blue}{(+0.8)}    & 92.6\textcolor{blue}{(+1.5)}    & 89.5\textcolor{blue}{(+2.2)} \\
    \textbf{Ours} (DETR) & 40 & 0.069 & 41,2 & 36.7\textcolor{blue}{(+1.0)}    & 77.1\textcolor{blue}{(+1.3)}    & 31.0\textcolor{blue}{(+2.6)} & 88.5\textcolor{blue}{(+0.3)}    & 82.4\textcolor{blue}{(+0.9)} \\
    \textbf{Ours} (Sparse R-CNN) & 34 & 0.073 & 48,4 & 37.3\textcolor{blue}{(+0.8)} & 73.4\textcolor{blue}{(+1.1)} & 34.2\textcolor{blue}{(+1.1)}    & 89.0    & 80.4\textcolor{blue}{(+0.7)} \\
    \bottomrule
    \end{tabular}%
    }
\end{minipage}
\vspace{0.5cm}


{\textbf{Setting of $\boldsymbol \lambda_{\boldsymbol1}$ , $\boldsymbol \lambda_{\boldsymbol2}$ , $\boldsymbol \lambda_{\boldsymbol3}$.} The total loss function of our method can be expressed as a weighted 
sum of $\mathcal{L}_{\text {class}}$,  $\mathcal{L}_{\text {box}}$ and $\mathcal{L}_{\text {clus}}$: 
\begin{equation}
\mathcal{L}=\lambda_{1} \mathcal{L}_{\text {clus}}+\lambda_{2} \mathcal{L}_{class}+\lambda_{3} \mathcal{L}_{box}
\end{equation}
Among them, $\lambda_{2}$ and $\lambda_{3}$ are set based on the default parameters of the original model, while $\lambda_{1}$ is set according to the rule that the loss we add is of the same order of magnitude as the other losses.
We applied our method on the Faster R-CNN, DETR and Sparse R-CNN models with different setting of $\lambda_{1}$, $\lambda_{2}$, $\lambda_{3}$, and the results obtained were shown in Table \ref{san}.}

\vspace{0.5cm}
\noindent
\begin{minipage}{\textwidth}
    \centering
    \captionof{table}{Effect of different setting of $\lambda_{1}$, $\lambda_{2}$, $\lambda_{3}$ on different models. The performance of None means the baseline performance.}
    \label{san}
    
    \resizebox{\textwidth}{!}{%
    \begin{tabular}{c cccc c cccc c ccc}
    \toprule
    \multirow{2}{*}{$\lambda_{1}$, $\lambda_{2}$, $\lambda_{3}$} & \multicolumn{3}{c} {Ours (Faster R-CNN)} & \multirow{2}{*}{$\lambda_{1}$, $\lambda_{2}$, $\lambda_{3}$} & \multicolumn{3}{c}{Ours (DETR)} & \multirow{2}{*}{$\lambda_{1}$, $\lambda_{2}$, $\lambda_{3}$} & \multicolumn{3}{c}{Ours (Sparse R-CNN)} \\
    \cmidrule(lr){2-4} \cmidrule(lr){6-8} \cmidrule(lr){10-12}
     & AP & AP$_{50}$ & AP$_{75}$ & & AP & AP$_{50}$ & AP$_{75}$ & & AP & AP$_{50}$ & AP$_{75}$ \\
    \midrule
      &47.0& 83.7 & 47.9 & 
        & 35.7 & 75.8 & 28.4& 
        & 36.5 & 72.3 & 33.1
        \\
    1,1,1  &48.1\textcolor{blue}{(+1.1)}& 86.4\textcolor{blue}{(+2.7)} & 48.7\textcolor{blue}{(+0.8)} & 1,5,1
        & 36.7\textcolor{blue}{(+1.0)} & 77.0\textcolor{blue}{(+1.2)} & 30.7\textcolor{blue}{(+2.3)}& 0.1,1,1
        & 37.2\textcolor{blue}{(+0.7)} & 72.4\textcolor{blue}{(+0.1)} & 34.6\textcolor{blue}{(+1.5)}
        \\
    0.33,0.33,0.33  &47.5\textcolor{blue}{(+0.5)}& 85.8\textcolor{blue}{(+2.1)} & 48.5\textcolor{blue}{(+0.6)} & 0.14,0.71,0.14
        & 34.8\textcolor{red}{(-0.9)} & 75.2\textcolor{red}{(-0.6)} & 28.3\textcolor{red}{(-0.1)}& 0.05,0.48,0.48
        & 36.8\textcolor{blue}{(+0.3)} & 72.1\textcolor{red}{(-0.3)} & 34.2\textcolor{blue}{(+0.9)}
        \\
    2,1,1  &47.2\textcolor{blue}{(+0.2)}& 85.4\textcolor{blue}{(+1.7)} & 48.3\textcolor{blue}{(+0.4)} & 2,5,1
        & 35.0\textcolor{red}{(-0.7)} & 75.0\textcolor{red}{(-0.8)} & 30.1\textcolor{blue}{(+1.6)}& 0.2,1,1
        & 36.2\textcolor{red}{(-0.3)} & 72.0\textcolor{red}{(-0.3)} & 34.0\textcolor{blue}{(+0.9)}
        \\    
    \bottomrule
  \end{tabular}%
    }
\end{minipage}
\vspace{0.5cm}

{As shown in Table \ref{san}, We attempted to set the sum of $\lambda_{1}$, $\lambda_{2}$, $\lambda_{3}$ to equal 1, or keep the values of $\lambda_{2}$ , $\lambda_{3}$ and only increase $\lambda_{1}$. However, the results were not better than keeping $\lambda_{2}$ and $\lambda_{3}$ in the original model set and then ensuring that $\lambda_{1}$ is of the same order of magnitude as them.}

{\textbf{Setting of clustering technique.} We applied our method on the Faster R-CNN, DETR and Sparse R-CNN models with different setting of clustering technique, namely {K-Means} and DBSCAN, and the results obtained were shown in Table \ref{DB}.}

\vspace{0.5cm}
\noindent
\begin{minipage}{\textwidth}
    \centering
    \captionof{table}{Effect of setting of clustering technique on different models. We use the parameters eps=12, min samples=5 and eps=5, min samples=2 and eps=5, min samples=2 for DBSCAN respectively. The performance of the None means the baseline performance.}
    \label{DB}
    
    \resizebox{\textwidth}{!}{%
    \begin{tabular}{c cccc ccc ccc}
        \toprule
        \multirow{2}{*}{} & \multicolumn{3}{c}{Faster R-CNN} & \multicolumn{3}{c}{DETR} & \multicolumn{3}{c}{Sparse R-CNN} \\
        \cmidrule(lr){2-4} \cmidrule(lr){5-7} \cmidrule(lr){8-10}
         & AP & AP$_{50}$ & AP$_{75}$ & AP & AP$_{50}$ & AP$_{75}$ & AP & AP$_{50}$ & AP$_{75}$ \\
        \midrule
        None   & 47.0 & 83.7 & 47.9 & 35.7 & 75.8 & 28.4 & 36.5 & 72.3 & 33.1 \\
        K-Means  & 48.1\textcolor{blue}{(+1.1)} & 86.4\textcolor{blue}{(+2.7)} & 48.7\textcolor{blue}{(+0.8)}
            & 36.0\textcolor{blue}{(+0.3)} & 76.7\textcolor{blue}{(+0.9)} & 28.9\textcolor{blue}{(+0.5)}
            & 37.2\textcolor{blue}{(+0.7)} & 73.4\textcolor{blue}{(+1.1)} & 34.1\textcolor{blue}{(+1.0)} \\
        DBSCAN  & 46.5\textcolor{red}{(-0.5)} & 83.3\textcolor{red}{(-0.4)} & 47.8\textcolor{red}{(-0.1)}
            & 30.5\textcolor{red}{(-5.2)} & 70.4\textcolor{red}{(-5.4)} & 21.7\textcolor{red}{(-6.7)}
            & 36.9\textcolor{blue}{(+0.4)} & 72.2\textcolor{red}{(-0.1)} & 34.1\textcolor{blue}{(+1.0)} \\
        \bottomrule
    \end{tabular}%
    }
\end{minipage}
\vspace{0.5cm}

{As shown in Table \ref{DB}, after changing the clustering method to DBSCAN, the detection effect of the model has decreased. The reason might be that: DBSCAN marks the points in the low-density area as noise, and our method precisely requires these points determined as noise as negative samples in contrastive learning.}

{\textbf{The reproducibility of result.}
In our method, we use random initialization for K-means. Considering the potential sensitivity of K-means to initialization, we conducted three experiments respectively on three models with the same parameters, and the results are as shown in Table \ref{re}.}

\vspace{0.5cm}
\noindent
\begin{minipage}{\textwidth}
    \centering
    \captionof{table}{{Effect of different random initializations on different models. We repeated 5 times with different random
initializations on different models.}}
    \label{re}
    
    \resizebox{\textwidth}{!}{%
    \begin{tabular}{p{13.465em}ccccc}
    \toprule
    Model & AP & AP$_{50}$ & AP$_{75}$ & AR$_{50}$ & F1-Score$_{50}$ \\
    \midrule
    Ours (Faster R-CNN) & $48.1\pm0.1$ & $86.0\pm 0.2$ & $49.6\pm 1.1$    & $91.8\pm 0.5$    & $88.8\pm 0.3$  \\
    Ours (DETR)  & $36.6\pm 0.2$ & $76.7\pm 0.2$ & $30.5\pm 0.2$    & $88.4\pm 0.3$    & $82.2\pm 0.2$  \\
    Ours (Sparse R-CNN) & $37.2\pm 0.1$ & $73.2\pm 0.1$ & $34.4\pm 0.2$    & $88.9\pm 0.1$    & $80.3\pm 0.1$  \\
    \bottomrule
    \end{tabular}%
    }
\end{minipage}
\vspace{0.5cm}

{As shown in Table \ref{re}, {each experiment was repeated 5 times with different random initializations. We report the mean
and standard deviation of the precision and recall.} The results are reproducible when rerunning the code. }

\section{Discussion}\label{sec6}
We propose a novel clustering-based feature space representation learning method to address the unique challenges of the detection of OBIs. The idea of our method is to guide the model to learn more discriminative feature
representations by utilizing the OBC font library as a clean,
expert-curated reference point. We conducted experiments on three main-stream detection frameworks: Faster R-CNN, DETR, and Sparse R-CNN on two OBIs detection datasets. Both numerical and visual experimental results prove that our method can improve the detection effect of the three models to some extent. {Our method offers an idea: how to use font libraries as the prior knowledge of the network to guide the learning of the network and improve the effect of the network, which is generalizable, particularly with the advancements of character databases and font libraries.}

\bmhead{Data availability}
The data are available at \url{https://jgw.aynu.edu.cn} and \url{https://www.jgwlbq.org.cn/home} respectively.
\bmhead{Code availability}
The code are available at \url{https://github.com/biscuit030/Clustering-based-Feature-Representation-Learning-for-Oracle-Bone-Inscriptions-Detection}
\bmhead{Acknowledgements}
This work is supported by the Young Scientists Fund of the National Natural 
Science Foundation of China (Grant No.62206106), and is a phased achievement of the National Social Science Foundation project "Research on Chinese Pre Qin Language and Culture Based on Jinwen Data" (No. 23VRC033), and has also received funding from the interdisciplinary cultivation project for young teachers and students at Jilin University, "Research on Jinwen Data Based on Artificial Intelligence" (No. 2024-JCXK-04), and is funded by the "Ancient Chinese Script and Chinese Civilization Inheritance and Development Project" under the project "Construction of Ancient Chinese Script Artificial Intelligence Recognition System" (Project No. G3829), and is supported by Graduate Work Department of Jilin University.
\bmhead{Author contributions}
XY developed the research idea and provided valuable suggestions for this 
manuscript. YT conducted the experiments in this manuscript. YT, HP, and XF
wrote this manuscript. CL provided specialized knowledge of ancient writing.
\bmhead{Competing Interests}
The authors declare no competing interests.

\small\bibliography{sn-bibliography}
\end{document}